\documentclass[10pt,twocolumn,letterpaper]{article}

\usepackage[pagebackref=true,breaklinks=true,letterpaper=true,colorlinks,bookmarks=false]{hyperref}
\usepackage{cvpr}
\usepackage{times}
\usepackage{titling}
\usepackage{epsfig}
\usepackage{graphicx}
\usepackage{amsmath}
\usepackage{amssymb}
\usepackage{multirow}
\usepackage{adjustbox}
\usepackage[table,xcdraw]{xcolor}

\newcommand*\samethanks[1][\value{footnote}]{\footnotemark[#1]}

\makeatletter
\newsavebox\saved@arstrutbox
\newcommand*{\setarstrut}[1]{%
  \noalign{%
    \begingroup
      \global\setbox\saved@arstrutbox\copy\@arstrutbox
      #1%
      \global\setbox\@arstrutbox\hbox{%
        \vrule \@height\arraystretch\ht\strutbox
               \@depth\arraystretch \dp\strutbox
               \@width\z@
      }%
    \endgroup
  }%
}
\newcommand*{\restorearstrut}{%
  \noalign{%
    \global\setbox\@arstrutbox\copy\saved@arstrutbox
  }%
}
\makeatother

\DeclareMathOperator{\EX}{\mathbb{E}}

\cvprfinalcopy 

\def\cvprPaperID{1920} 

\ifcvprfinal\fi

\begin{document}

\title{Pay attention! - Robustifying a Deep Visuomotor Policy through Task-Focused Visual Attention}
\author{Pooya Abolghasemi\thanks{Authors contributed equally.}, Amir Mazaheri\samethanks, Mubarak Shah and Ladislau B\"ol\"oni\\
University of Central Florida, Orlando, FL 32816\\
{\tt\small pooya.abolghasemi, amirmazaheri@knights.ucf.edu, shah@crcv.ucf.edu, lboloni@cs.ucf.edu}
}
\date{}

\maketitle

\begin{abstract}
Several recent studies have demonstrated the promise of deep visuomotor policies for robot manipulator control. Despite impressive progress, these systems are known to be vulnerable to physical disturbances, such as accidental or adversarial bumps that make them drop the manipulated object. They also tend to be distracted by visual disturbances such as objects moving in the robot's field of view, even if the disturbance does not physically prevent the execution of the task. In this paper, we propose an approach for augmenting a deep visuomotor policy trained through demonstrations with Task Focused visual Attention (TFA). The manipulation task is specified with a natural language text such as ``move the red bowl to the left''. This allows the visual attention component to concentrate on the current object that the robot needs to manipulate. We show that even in benign environments, the TFA allows the policy to consistently outperform a variant with no attention mechanism. More importantly, the new policy is significantly more robust: it regularly recovers from severe physical disturbances (such as bumps causing it to drop the object) from which the baseline policy, i.e. with no visual attention, almost never recovers. In addition, we show that the proposed policy performs correctly in the presence of a wide class of visual disturbances, exhibiting a behavior reminiscent of human selective visual attention experiments. Our proposed approach consists of a VAE-GAN network which encodes the visual input and feeds it to a Motor network that moves the robot joints. Also, our approach benefits from a teacher network for the TFA that leverages textual input command to robustify the visual encoder against various types of disturbances.

\end{abstract}

\section{Introduction}

\begin{figure}
    \centering
    \includegraphics[width=1.0\columnwidth]{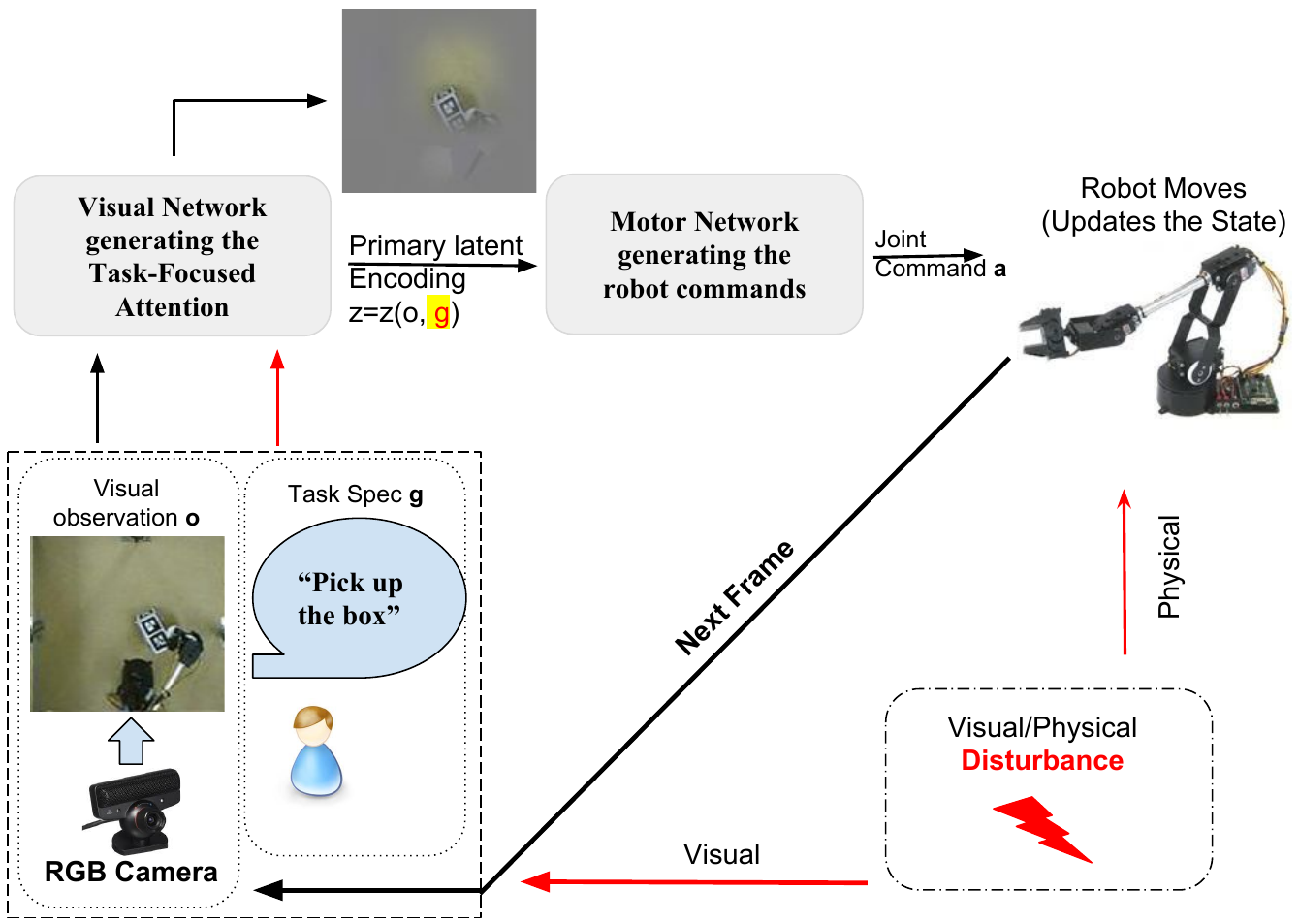}
    \caption{The robot performs a given command. Our proposed network attends the image regions that matter the most, and is robust to physical and visual \em{disturbance}.}
    \label{fig:Intro}
\end{figure}
Many recent researches show the possibility of end-to-end training of deep visuomotor policies that perform object manipulation tasks such as pick-and-place, push-to-location, stacking and pouring. These systems perform all the components of the task (vision processing, grasp and trajectory planning and robot control) using a neural network trained by variations of deep reinforcement learning and learning from demonstration (supervised learning). 


Deep visuomotor policies for manipulator control are neural network architectures that have as input an observation composed of an image or video frame and possibly other sensor data, $\mathbf{o}_t$, a task (or goal) specification, $\mathbf{g}$,  and output robot commands, $\mathbf{a}_t=\pi(\mathbf{o}_t,\mathbf{g})$. The robot executes these commands, enacting a change in the external environment, which creates a new observation $\mathbf{o}_{t+1}$, and the cycle repeats. Architecturally, most currently proposed systems follow variations of the generic model of Figure~\ref{fig:Intro}, which posits the existence of a primary latent encoding, $\mathbf{z}$, the result of the visual processing of the input by a specialized visual network. This encoding, of dimensionality orders of magnitude smaller than the input, is then used by the motor network to generate the next state joint angles command, $\mathbf{a}$.  
While most demonstrations (supervised data) had been made in unstructured but relatively benign environments, our own experiments and personal communication with other researchers had shown that task independent visual networks for visuomotor policies are highly vulnerable to {\em physical} and {\em visual} disturbances. An example of physical disturbance is the robot arm being bumped such that it drops the manipulated object. The desired behavior would be for the robot to immediately notice this, change its trajectory, pick up the dropped object and continue with the manipulation task. Instead, with an otherwise reliably performing policy, we notice situations where the robot arm, having lost the object, continue to go empty-handed through the full trajectory of the manipulation, recovering either much later, or not at all. A visual disturbance may involve distracting mobile objects appearing in the robot's field of view. Clearly, if the visual disturbance prevents the execution of the task, for instance, by blocking the view of the manipulated object, it is acceptable for the robot to stop or even cancel the manipulation. There are, however, visual disturbances that should not prevent the execution of the task: for instance, hands waving in the visual field of the robot but not covering the manipulated object or the robot arm. We have found that in the case of an task independent visual network, even such visual disturbances cause the robot to behave erratically -- possibly due to the robot interpreting the situation as a state never encountered before. 

In engineered robot architectures such problems can be dealt by developing explicit models of the possible disturbances, which may allow the robot to reason around the situation. In deep learning systems, one possible brute-force solution is to gather more training data containing physical and visual disturbance events; however, data collection for robotic tasks is time consuming. Also, there are unlimited visual and physical disturbance scenarios for a single task. It is impossible for to record demonstrations to cover all possible scenarios of physical and visual disturbances.

\begin{figure*}[ht]
    \centering
    \includegraphics[width=\textwidth]{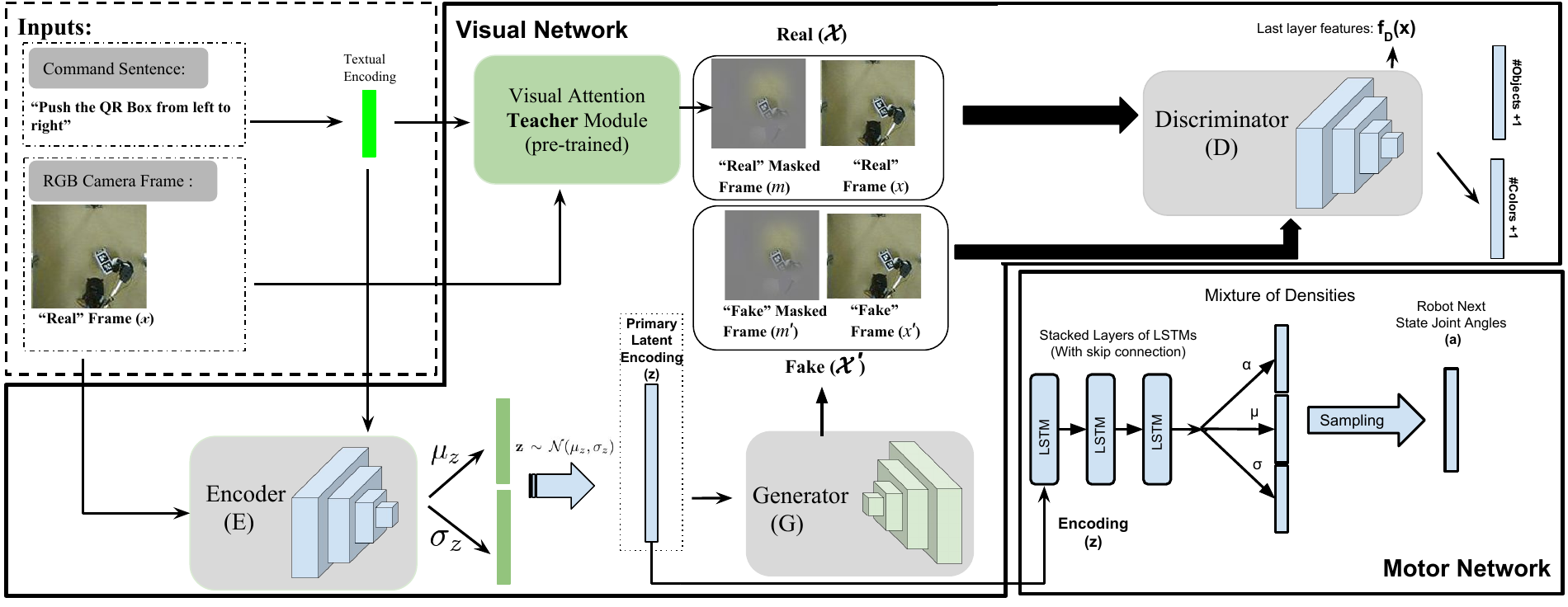}
    \caption{\textbf{The proposed visuomotor architecture}. Given an image captured from the scene and a command sentence provided by the user, the \textbf{Encoder (E)} produces the \textit{Primary Latent Encoding (z)}.  \textbf{z} is the input to the \textbf{Motor Network}, which decides the next state of the robot joint angles. Also, \textbf{z} is the input to a \textbf{Generator (G)}, which produces ``Fake'' frame and masked frame. A pre-trained Visual Attention \textbf{Teacher} Module masks the original frame by an spatial attention computed employing the textual input. The \textbf{Discriminator (D)} must discriminate between real/fake frames and masked frames, and also classify the object and color of the object being manipulated.}
    \label{fig:VisuoMotorArchitecture}
\end{figure*}

\paragraph{Pay attention! Task dependent visual network:} The principal idea of this paper is that performance benefits can be obtained if we make the vision system pay attention to relevant regions of each frame regarding the current task or user command. Humans are known to exhibit selective attention - when observing a scene with a particular task in mind, features of the scene relevant to the task are given particular attention, while other features are de-emphasized or even ignored. This had been illustrated in the famous experiments of Chabris and Simmons~\cite{chabris2010invisible}. In this paper we propose {\em Task Focused (Visual) Attention} (TFA) as an auxiliary network to increase the robustness of the robot manipulator network to physical and visual disturbances, without the need of any additional training data. Thus, our objective is  to create a system that implements a selective visual attention similar to what human perception is doing: we want the robot to focus on the objects of the scene that are relevant to the current manipulation task. We conjecture that using TFA, $\mathbf{z}$ will better represent the objects and colors that are the subject of the attention, allowing for more precision in grasping and manipulation (See Figure~\ref{fig:VisuoMotorArchitecture}).

\paragraph{Our Contributions:} The contributions of the paper are as follows:
\textbf{1-} We describe a novel architecture for a visuomotor policy trained end-to-end from demonstrations, which features a task focused visual attention system. The visual attention system is guided by a natural language description of the task and focuses on the currently manipulated object. \textbf{2-} We show that, under benign conditions, the new policy outperforms a closely related baseline policy without the attention model over pick-up and push tasks using a variety of objects. \textbf{3-} We show that in the case of a severe physical disturbance, when an external intervention causes the robot to miss the grasp or drop the already grasped object, the new policy recovers in the majority of situations, while the baseline policy almost never recovers. \textbf{4-} We show that the task focused visual attention allows the policy to ignore a large class of visual disturbances, that interfere with the task for the baseline policy. We show experimentally that the system exhibits the ``invisible gorilla'' phenomenon~\cite{chabris2010invisible} from the classic selective attention test. \textbf{5-} The teacher network for the task focused visual attention can be trained offline, does not require additional training data or pixel level annotation of objects.

\section{Related Work}
A deep visuomotor policy for robotic manipulation transforms an input video stream (possibly combined with other sensory input) into robot commands by the means of a single deep neural network. Such a system had been first demonstrated in~\cite{levine2016end} using guided policy search, a method that transforms policy search into supervised learning, with supervision provided by a trajectory-centric reinforcement learning method. In recent years, several alternative approaches have been proposed using variations of both deep reinforcement learning and deep learning from demonstration (as well as combinations of these). 

Deep reinforcement learning is powerful paradigm which, in applications where exploration can be performed in a simulated environment allowing millions of trial runs, can train systems that perform at superhuman level~\cite{silver2016mastering}, even when no human knowledge is used for bootstrapping~\cite{silver2017mastering}. Unfortunately, for training visuomotor policies controlling real robots, it is very difficult to perform reinforcement runs on these scales. Even the most extensive projects could only collect several orders of magnitude lower number of experiments: for example, in~\cite{levine2018learning} 14 robotic manipulators were used over the period of two months to gather 800,000 grasp attempts. Even this number of experimental tries are unrealistic in many practical settings. 

Thus, many efforts focus on reducing the number of experimental runs necessary to train an end-to-end visuomotor controller. One obvious direction is to learn a better encoding of the input data, which can improve the learning rate. In~\cite{finn2015deep}, a set of visual features were extracted from the image to be used as state representation for a reinforcement learning algorithm. 

Another direction involves the use of learning from demonstration instead (or in combination with) of reinforcement learning. The demonstrations can be performed in real~\cite{rahmatizadeh2017vision} or simulated~\cite{james2017transferring,zhu2018reinforcement} environments. Meta-learning~\cite{duan2017one} and related approaches promise to drastically lower the amount of training data needed to learn a specific task from a class of related tasks (possibly, down to a single task specific demonstration). However, they still require a costly meta-learning phase. 


An approach that is similar to ours in objective, but different in implementation, is described in~\cite{devin2017deep}. Considering manipulation tasks, the authors implement two layers of attention. The first, a task independent visual attention  semantically identifies labels and localizes objects in the scene. This labeling relies on training on an external labeled dataset, thus in this respect the approach is not ``end-to-end''. The second, a task-specific attention is learned by selecting from the segmented objects,  by the task independent attention, those objects that contribute most to the correct prediction of demonstrated trajectories. 




Another point concerns the way in which the task is specified to the robot. Specifying the task in the form of a human readable sentence is a natural choice~\cite{tellex2011understanding}, as creating such a command is very easy for a human user. In the general case, however, translating a command into a task is not yet feasible with an end-to-end learned controller. In this paper, we assume the existence of the command, but only as an additional input that helps the creation of the task-focused attention. Alternative ways of specifying the task are possible. A purely visual specification was proposed in~\cite{finn2017deep}, where the user identifies a pixel in the image and specifies where it should be moved. A technique of control based on visual images was also demonstrated in~\cite{watter2015embed}.

\medskip

One component of our work has its roots in recent work on visual attention networks. These networks often appear as components of larger networks, solving problems like image captioning~\cite{xu2015show,you2016image}, visual question answering~\cite{yang2016stacked,Mazaheri_2017_ICCV,yu2017multi} or visual expression localization~\cite{yu2018mattnet}. Although the applications are different, the role of attention networks, i.e., focusing on information-rich parts of the visual input, remains the same. Our proposed attention mechanism is most similar to~\cite{yang2016stacked}. However, in our model we train the attention network with word selection objective. The objective is to select some regions on a video frame regarding a textual input, such that it be able to regenerate the words in the input sentence just based on the visual features of selected image regions.

\section{Approach}

As shown in Figure~\ref{fig:VisuoMotorArchitecture}, our architecture contains a \textbf{Motor Network} and \textbf{Visual Network}.

The Motor Network, often but not always, contains a recurrent neural network and is trained on a loss that favors the execution of the specified task, $\mathbf{g}$. This training may take several forms. In the case of RL we need a source of rewards. If the task is specified by demonstrations (our case), the training may be executed in a supervised fashion using a behavioral cloning loss. 

The Visual Network contains an Encoder module that encodes the input frame into the \textit{Primary Latent Variable, \textbf{z}}. To get a richer representation \textbf{z}, we incorporate two other modules. First, a teacher network  which computes an attention map and masks the input frame. We train the teacher network separately (Section~\ref{sec:attention}).  Second, a GAN network that takes \textbf{z} as input and generates two reconstructed frames, the input frame and the masked input frame.

\subsection{A Teacher Network for TFA}
\label{sec:attention}

\begin{figure}
    \centering
    \includegraphics[width=0.85\columnwidth]{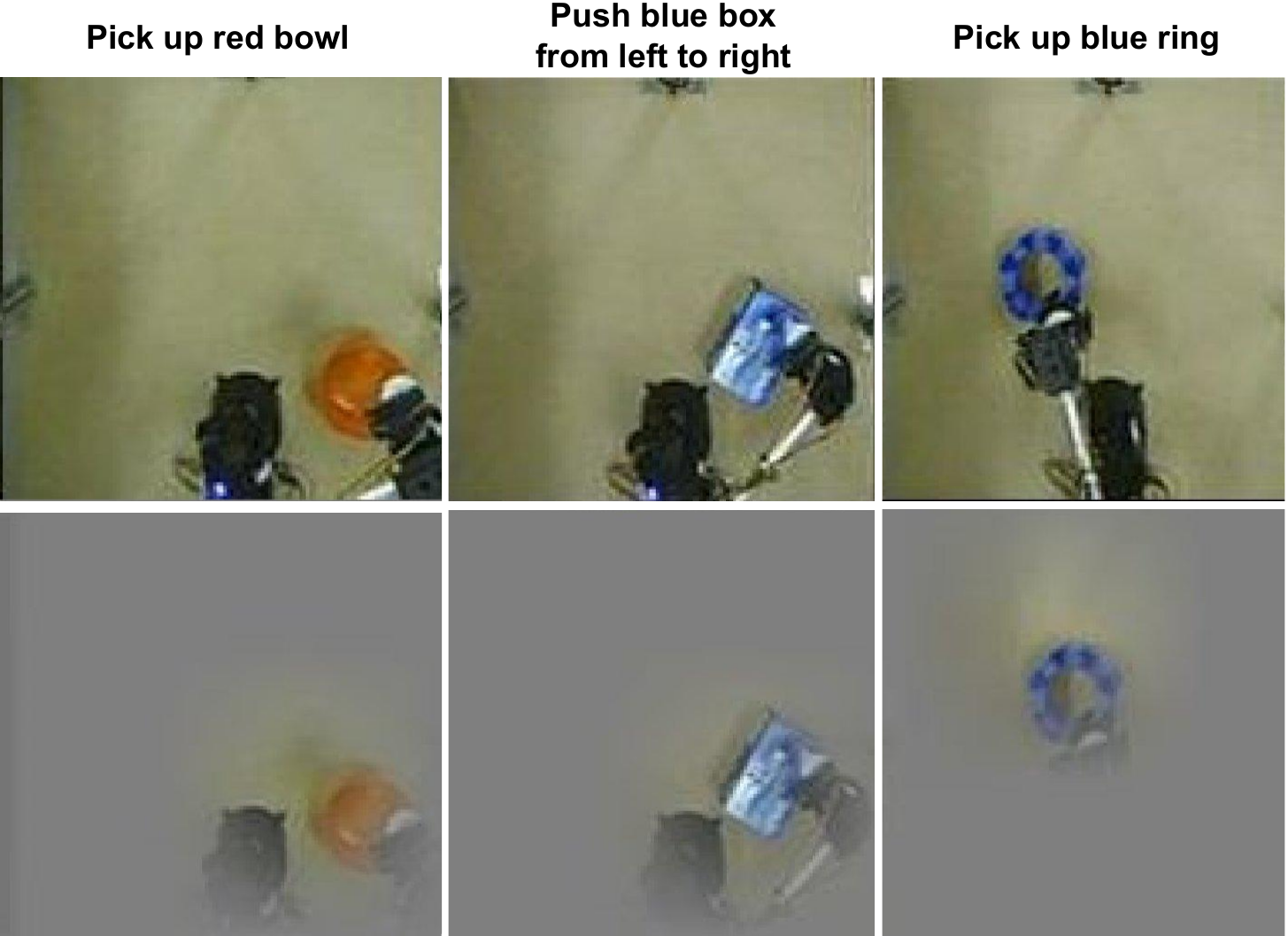}
    \caption{Examples of task focused visual attention. 
    We provide the command sentence on top of each column. The first row shows frames from RGB camera and the second row is the same image masked by the attention, produced by \textbf{teacher network}. We denote the first/second row images by $x$/$m$ in our equations.}
    \label{fig:TFA-example}
\end{figure}

We  consider robot manipulation commands expressed in natural language such as, ``Push the {\em red plate} to the left'', ``Push the {\em blue box} to the left'', and ``Pick up the {\em red ring}''.

The goal of the TFA is to identify the parts of the visual input, where objects relevant to the task appear, that is, to focus the attention on the red plate, blue box and blue ring respectively (see Figure~\ref{fig:TFA-example}). 

A TFA system could be trained as a supervised learning model, if we can create a sufficient amount of training data. However, this would require us to label with attention blobs on an unrealistically large number of input video frames. Our approach is to generate our own labels by implementing a teacher network that provides training data for the controller. Our approach fits in the established technique of student-teacher network training models~\cite{lawrence1997lessons,bucilua2006model,hinton2015distilling}, with the qualification that the attention teacher only teaches one particular aspect of the final controller. 

\begin{figure}
    \centering
    \includegraphics[width=\columnwidth]{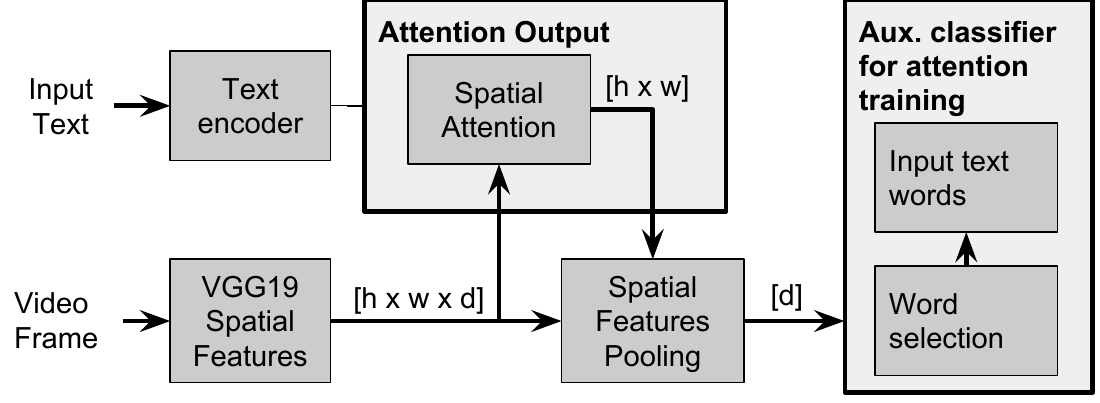}
    \caption{Proposed visual attention network. The network uses the pre-trained VGG19~\cite{simonyan2014very} network's last convolution layer output as the visual spatial features. The attention module combines the spatial and textual features, and assigns one probability to each spatial region. To train the attention network, first we pool the visual features by the attention probabilities (weighted average), and second, we use an auxiliary classifier to reconstruct the input text's words based on the pooled visual features.}
    \label{fig:AttFlow}
    \vspace{-5mm}
\end{figure}

In the remainder of this section, we describe the implementation of a teacher network which computes the TFA as shown in Figure~\ref{fig:AttFlow}. 

The proposed approach allows us to train the TFA without pixel level annotations. The principal idea is that the attention should be on those regions that {\em allow us to reconstruct the input text based on those regions only}. The overall architecture is described in Figure~\ref{fig:AttFlow}.

We divided the visual field (video frame), $x$, into $k$ regions. The visual attention we aim to obtain is a vector of probabilities, $p_\textit{TFA} \in (0,1)^{k}$, with a probability for each of the $k$ regions. The higher the probability, the more attention is paid to the specific region. In general, our goal is to focus the attention on a small number of regions. 


The first step is to encode the text and image inputs. 

\medskip

\noindent {\bf Text input:} Let $\{v_1, v_2, \dots, v_n\}$ be the textual input with $n$ words, with one-hot indicators $v_i \in \{0,1\}^{|V|}$, where $V$ is the dictionary of the words in our dataset. One-hot vectors are insufficient and redundant representations. Thus, a word-to-vector encoding is employed: 
\begin{equation}
    w_i = v_i \times W_\omega,
\end{equation}
where $W_\omega \in \mathcal{R}^{|V| \times d_v}$, and $d_v$ is the length of encoded word vectors. To encode a whole sentence, we feed the series of word vectors to an LSTM. To obtain the text encoding, we extract the last hidden state of the LSTM, $u \in \mathcal{R}^{d_h}$, where $d_h$ is cell size of the LSTM.

\medskip

\noindent {\bf Visual input:} To obtain the visual encoding, we divide the visual input (video frame)  into $k$ spatial regions and individually process them to extract visual features using a VGG19~\cite{simonyan2014very} network. The resulting spatial visual features  have the form $\phi_{f} \in \mathcal{R}^{k \times d_\phi}$, where $k$ is the number of spatial regions and $d_\phi$ is the length of feature vector for each region.


We combine the textual and visual encodings through a technique similar to \cite{yang2016stacked}. We learn a mapping on both visual and text data and combine them through an element-wise summation:
\begin{equation}
    \psi = \tanh(\phi_f \times W_f \oplus u \times W_u),
\end{equation}
where $W_u \in \mathcal{R}^{d_h \times d_\psi}$ and $W_f \in \mathcal{R}^{d_\phi \times d_\psi}$ are mapping matrices, $\oplus$ is element-wise summation. $\psi \in \mathcal{R}^{k \times d_\psi}$ is the combination matrix of textual and visual inputs. Note that, $u$ is a vector, while $\phi_f$ is a matrix. We augment the $u$ vector by repeating it for $k$ times. To compute the final attention probabilities, the model must assign higher scores to a few spatial regions.
\begin{equation}
    p_\textit{TFA} = \mathrm{softmax}(\psi \times W_p),
\end{equation}
where $W_p \in \mathcal{R}^{d_{\psi} \times 1}$ is trainable weights vector, which is used to assign a score to each region. The final $p_\textit{TFA} \in (0,1)^{k}$ is the vector containing attention scores of all $k$ regions'. We use a $\mathrm{softmax}$ non-linearity to force the network to attend to a few number of regions.

Our method does not need any spatial pixel level annotation to compute the attention. The attention in our formulation is a latent variable dependent on the input text and frame (See Figure~\ref{fig:AttFlow}). The main idea which allows us to train the attention network, is that from a pooled spatial features weighted by the latent variable attention, $p_\textit{TFA}$,  we should be able to reconstruct the input text (user command sentence) words  $\mathcal{V} \in \{0,1\}^{|V|}$. Here, we define the weighted pooled features $u \in \mathcal{R}^{d_{\phi}}$:
\begin{equation}
    u = \sum_{i \in k}{p_{\textit{TFA}_i} {\phi_f}_i }.
\end{equation}

Basically, given a video frame and sentence, we force the network to select a few regions of the input frame, and reconstruct the input text just based on the selected regions. As a result, the only way that the network can reconstruct the original input text, is by selecting the relevant regions of the frame:
\begin{equation}
    \hat{\mathcal{V}} = \sigma( \tau (u)),
\end{equation}
\noindent where $\tau(.)$ is a multi-layer perceptron. $\hat{\mathcal{V}} \in (0,1)^{|V|}$ contains the predicted set of words. We optimize the entropy loss function $\mathcal{L}_{att} = - \mathcal{V}log(\hat{\mathcal{V}})$.

In Figure~\ref{fig:TFA-example}, we show RGB frames and the masked frame using computed attention $p_{\textit{TFA}}$. To mask the RGB frames, we reshape and re-size $p_{\textit{TFA}}$ (using bi-linear interpolation) to the same size of the input frame (x); followed by smoothing the mask by applying a Gaussian filter on it. We denote the masked RGB input frame with the attention $p_{\textit{TFA}}$ by $m$.
\subsection{The visual and motor networks}
\label{sec:vismotnetworks}
Our architecture follows the generic architecture for the visuomotor policy in Figure~\ref{fig:Intro}. It consists of a {\bf Visual Network} sub-module that extracts a primary latent encoding, $\mathbf{z}$, and a {\bf Motor Network} that transforms $\mathbf{z}$ into actions, which in our case are joint angle commands (next state of the robot arms). However, our architecture makes several specific decisions with the aim to take advantage of the available the text description of the current task and the TFA. 

\subsubsection{Visual Network}
\label{sec:visualNetwork}
The objective of the Visual Network is to create a compact primary latent encoding that captures the important aspects of the current task. An ongoing problem is that the encoding needs to work within a certain limited dimensionality budget. Intuitively, general purpose visual features extracted from the image would waste space by encoding aspects of the image that are not relevant to the task. On the other hand, focusing only on the attention field may ignore parts of the image that are important for the task. For instance, in Figure~\ref{fig:TFA-example}- bottom right masked frame, the robot arm itself is not visible. 

Our proposed architecture for the visual network,   shown in  Figure~\ref{fig:VisuoMotorArchitecture},   incorporates several techniques that allows it to learn a representation that efficiently encodes the parts of the input that are relevant to the {\em current task}. The overall architecture follows the idea of a VAE-GAN~\cite{larsen2015autoencoding}: it is composed of an encoder, a generator and a discriminator. The {\em Primary Latent Encoding ($\mathbf{z}$)} is extracted from the output of the visual encoder (E).

The visual network receives a raw frame $\mathbf{x}$ and a one-hot representation of the user command (input sentence), denoted by $I_c \in \{0,1\}^{|V|}$. In fact, $I_c$ is indicates which words of the dictionary are appearing in the textual input command. We assume that $\mathbf{z} \sim \mathcal{N}(\mu_{z},\sigma_{z})$, and:
\begin{align}
[\mu_{\mathbf{z}} | \sigma_{\mathbf{z}}] = E(x, I_c),
\end{align}
where $\mu_{z}$, $\sigma_{z} \in \mathcal{R}^{d_z}$, and $d_z$ is the length of the Primary Latent Encoding  (z). In fact, $E$ is a multi-layer convolutional neural network with a $2d_z$ dimensional vector which splits into $\mu_z$ and $\sigma_z$.

The generator, (G), takes the Primary Latent Encoding $\mathbf{z}$ as input, and produces two images, a reconstruction frame, and a reconstructed frame masked with attention (``Fake Frame'' and ``Fake masked frame'' in Figure~\ref{fig:VisuoMotorArchitecture}). Notice that a novel aspect of our proposed architecture is that the generator does not only create a reconstruction of the input, $\mathbf{x}'$, but also an approximation of the faked masked frame, $m'$.

Unlike traditional GAN discriminators, the discriminator $D$ employed in our architecture performs a more complicated classification~\cite{salimans2016improved}. Masked and unmasked frames($m/m'$, $x/x'$) are both inputs to the discriminator, and it classifies the objects ($s$) and color ($c$) of the object of interest, as well as whether the input was fake or real. The discriminator has two outputs of lengths of  $|s| + 1$ and $|c| + 1$. $|s|$ and $|c|$ are respectively the number of colors and objects in the vocabulary $|V|$ and the ``$+1$'' is for the ``fake'' class. We make the set of $s$ and $c$ tags by parsing all the input sentences (user's textual commands) in the training. 
\subsubsection{Motor Network}

The motor network in our architecture (see Figure~\ref{fig:VisuoMotorArchitecture}) contains both recurrent and stochastic components. It takes as input the primary latent encoding, $\mathbf{z}$, which is processed through a 3-layer LSTM network with skip connections~\cite{graves2013generating}. Note that the memory cells of LSTMs get updated through the time by doing the task (frame by frame). The output of the final LSTM layer is fed into a mixture density network (MDN)~\cite{bishop1994mixture}. MDN provides a set of Gaussian kernels parameters namely $\mu_{i}$, $\sigma_{i}$ and the mixing probabilities $\alpha_{i}(x)$, all $ \in \mathcal{R}^{|J|}$, and $1 \leq i \leq N_G$. Here,  $|J|$ is the number of robot joints (specific to the robot) and $N_G$ is the number of Gaussian components. The $|J|$-dimensional vector describing the next joint angles is sampled from this mixture of Gaussians. 

We provide the detailed architectures of D, G, E, and motor sub-networks in the Supplementary Material.

\subsection{Loss Function and Training}
In this section, we describe the discriminator loss function $\mathcal{L}_{D}$, and the generator loss function $\mathcal{L}_{G}$. All the parameters in the Discriminator have been optimized to minimize $\mathcal{L}_{D}$, and parameters of the visual Encoder, Generator, and Motor network are optimized by the loss value $\mathcal{L}_{G}$ in a GAN training manner. In following, to prevent repetition of equations, we use the unifying tuples $\mathcal{X}' = (x', m')$ and $\mathcal{X} = (x, m)$ as fake and real data respectively. To clarify, $(x', m') = G(\mathbf{z} \sim E(x, I_c))$, while $x$ is the real frame from RGB camera, and $m$ is the masked real frame by the teacher network (Section~\ref{sec:attention}).

\subsubsection{Discriminator Loss}
If the discriminator $D$ is receiving real data $\mathcal{X}$, it needs to classify the object and color contained in the user's textual command input:
\begin{align}
    \mathcal{L}_\mathit{real} =& - \EX_{\mathcal{X}, s \sim p_\mathit{data}} [\log{(P_\mathit{D}(s \big\vert \mathcal{X}))}] \nonumber\\
           & - \EX_{\mathcal{X}, c \sim p_\mathit{data}} [\log{(P_\mathit{D}(c \big \vert \mathcal{X}))}],
\end{align}
where $P_{\mathit{D}}$ is the class probabilities produced by the discriminator for both colors and objects. Similarly, if $D$ receives $\mathcal{X}'$, it should classify them as fake:
\begin{align}
    \mathcal{L}_\mathit{fake} =& - \EX_{\mathcal{X}' \sim G} [\log{(P_\mathit{D}(|s|+1 \big \vert \mathcal{X}'))}] \nonumber \\
            &- \EX_{\mathcal{X}' \sim G} [\log{( P_\mathit{D}(|c|+1 \big \vert \mathcal{X}'))}].
\end{align}

Finally, if $D$ receives raw and masked faked frames, generated by $G$ with the latent representation $\mathbf{z} \sim \mathcal{N}(0, 1)$:
\noindent \begin{align}
    \mathcal{L}_\mathit{noise}\!=\!
    & - \EX_{z \sim \mathit{noise}} [\log{( P_\mathit{D}(|s|+1 \big \vert G(z)))}] \nonumber\\
            &- \EX_{z \sim \mathit{noise}} [\log{(P_{D}(|c|\!+\!1 \big \vert G(z)))}].
\end{align}

The overall loss of the discriminator is thus $\mathcal{L}_{D} = \mathcal{L}_\mathit{real} + \mathcal{L}_\mathit{fake} + \mathcal{L}_\mathit{noise}$.
\subsubsection{Generator Loss}
The Generator (G) must reconstruct a real looking frame and masked frame by attention that contain the object of interest. In fact, G tries not only to look real, but also presents the correct object in both of its outputs. Hence, it has to fool the discriminator which tries to distinguish between fake frames and different objects and colors:
\begin{align}
    \mathcal{L}_\mathit{GD} =& - \EX_{\mathcal{X}', s \sim p_G} [\log{p_\mathit{D}(s \big \vert \mathcal{X}')}] \nonumber\\
           & - \EX_{\mathcal{X}', c \sim p_G} [\log{p_\mathit{D}(c \big \vert \mathcal{X}')}].
\end{align}

The training of GANs is notoriously unstable. A possible technique to improve stability is {\em feature matching}~\cite{bao2017cvae}-- forcing $G$ to generate images that match the statistics of the real data. Here, we use features extracted by the last convolution layer of $D$ for this purpose and we call it $f_{D}(x)$. The generator must produce outputs that have similar $f_{D}$ representation to real data. We define the loss term $\mathcal{L}_\mathit{fea}$ as a distance between the real inputs $x$/$m$ and generated ones $x'$/$m'$ features~\cite{salimans2016improved}:
\begin{align}
    \mathcal{L}_\mathit{fea} &= ||f_{D}(x) - f_{D}(x')||^2 + ||f_{D}(m) - f_{D}(m')||^2.
\end{align}

To regularize the Primary Latent Encoding ($\mathbf{z}$), we minimize the KL-divergence between $\mathbf{z}$ and $\mathcal{N}(0,1)$:
\begin{align}
    \mathcal{L}_\mathit{prior} &= D_\mathrm{KL}(E(x, I_c)~||~\mathcal{N}(0, 1)).
\end{align}

Additionally, a reconstruction error of ``fake'' Frame/Masked generated by G is defined by:
\begin{equation}
\mathcal{L}_\mathit{rec} = ||x' - x||^2 + ||m' - m||^2.    
\end{equation}

\paragraph{Motor Network Loss:}The motor loss is calculated according to the MDN negative log-likelihood loss formula over the supervised data based on the demonstrations (behavioral cloning loss):
\begin{align}
    \mathcal{L}_\mathit{motor} = -log{\left(\sum_{i=1}^{N_G}\alpha_{i}(x)\cdot P_{\sim \mathcal{N}(\mu_i, \sigma_i)}(J) \right)}.
\end{align}

Finally, we write the Generator loss as $\mathcal{L}_{G} = \mathcal{L}_\mathit{DG} + \mathcal{L}_\mathit{rec} + \mathcal{L}_\mathit{prior} + \mathcal{L}_\mathit{motor}$.


\section{Experiments}
\label{sec:result}
\begin{figure*}[ht]
    \centering
    \includegraphics[width = 0.85\textwidth]{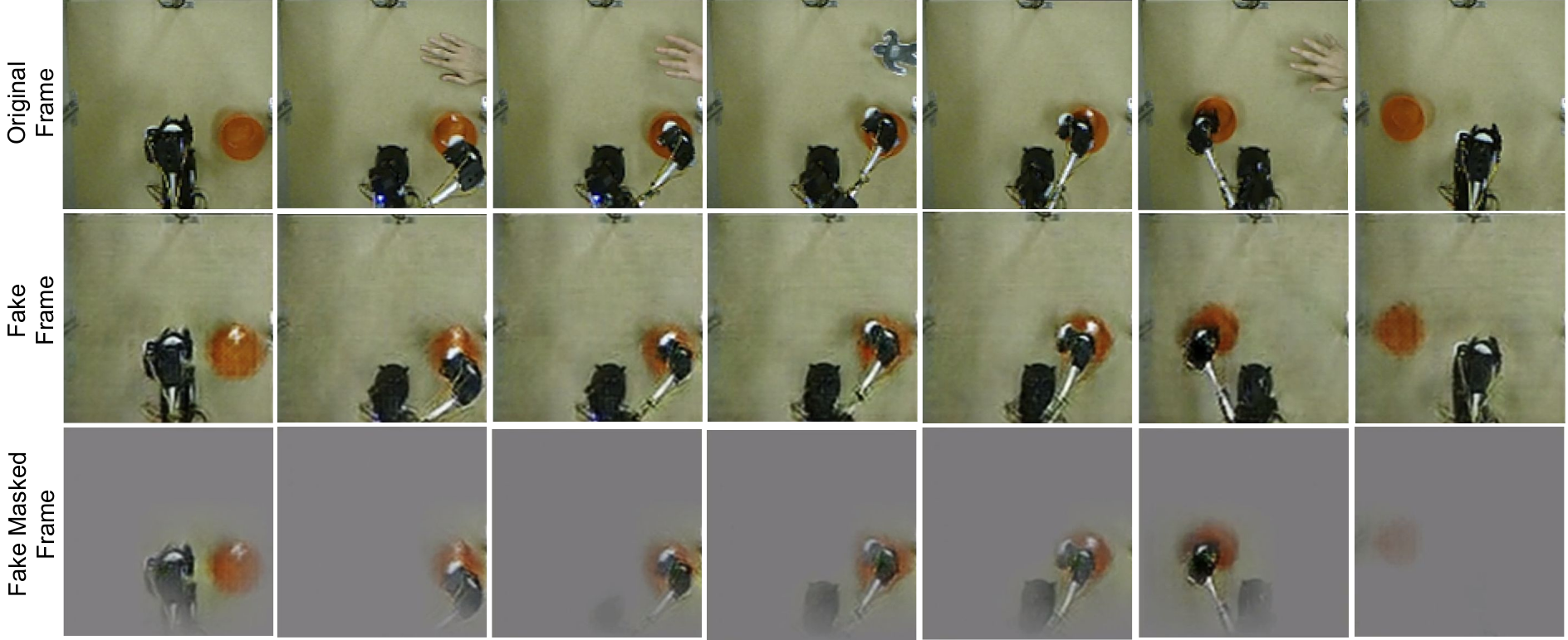}
    \caption{An execution of the pushing task with the sentence ``Push the red bowl from right to left". Top row: original input image, middle row: fake frame generated by the \textbf{Generator(G)}, bottom row: fake masked image with TFA generated by \textbf{G}. You can compare the fake masked frames presented in this figure with attention maps generated by the \textbf{teacher network} in Figure~\ref{fig:TFA-example}. Notice that visual disturbances such as the hand and the gorilla do not appear in the reconstructed image.}
    \label{fig:gorilla_effect}
\end{figure*}
We collected demonstrations for the tasks of picking up and pushing objects using an inexpensive Lynxmotion-AL5D robot. We controlled the robot using a PlayStation controller. For each task and object combination we collected 150 demonstrations. The training data consists of joint-angle commands plus the visual input recorded in 10 fps rate by a PlayStation Eye camera mounted over the work area. The training data thus collected was used to train both the Visual and the Motor Networks. Note that this robot does not have proprioception -- any collision or manipulation error needs to be detected solely from the visual input.

%

\subsection{Performance under benign conditions}
\begin{table*}[ht]
\begin{adjustbox}{width=\textwidth}
\begin{tabular}{p{2cm}ccccccccccccc}
\cline{1-13}
\multicolumn{1}{|c|}{\multirow{3}{*}{{\textbf{\begin{tabular}[c]{@{}c@{}}\\Textual\\Command\\ Sentences\\ \\ \\ \end{tabular}}}}} & \multicolumn{7}{c|}{\multirow{2}{*}{{\textbf{Pick up ...}}}} & \multicolumn{5}{c|}{\multirow{2}{*}{{\textbf{Push ... from left to right}}}} &  \\
\multicolumn{1}{|c|}{} & \multicolumn{7}{c|}{} & \multicolumn{5}{c|}{} & \multicolumn{1}{c}{} \\ \cline{2-14}
\multicolumn{1}{|c|}{} & \multicolumn{1}{c|}{\rotatebox[origin=c]{-90}{{\begin{tabular}[c]{@{}c@{}}Red \\ Bowl\end{tabular}}}} & \multicolumn{1}{c|}{\rotatebox[origin=c]{-90}{{\begin{tabular}[c]{@{}c@{}}White \\ Towel\end{tabular}}}} & \multicolumn{1}{c|}{\rotatebox[origin=c]{-90}{{\begin{tabular}[c]{@{}c@{}}Blue \\ Ring\end{tabular}}}} & \multicolumn{1}{c|}{\rotatebox[origin=c]{-90}{{\begin{tabular}[c]{@{  }c@{  }}Black \\Dumbbell \end{tabular}}}} & \multicolumn{1}{c|}{\rotatebox[origin=c]{-90}{{\begin{tabular}[c]{@{}c@{}}White \\ Plate\end{tabular}}}} & \multicolumn{1}{c|}{\rotatebox[origin=c]{-90}{{\begin{tabular}[c]{@{}c@{}}Red \\ Bubbles\end{tabular}}}} & \multicolumn{1}{c|}{\rotatebox[origin=c]{-90}{{\begin{tabular}[c]{@{}c@{}}Mean \\ pick up\end{tabular}}}} & \multicolumn{1}{c|}{\rotatebox[origin=c]{-90}{{\begin{tabular}[c]{@{}c@{}}Red \\ Bowl\end{tabular}}}} & \multicolumn{1}{l|}{\rotatebox[origin=c]{-90}{{\begin{tabular}[c]{@{}c@{}}White \\ Plate\end{tabular}}}} & \multicolumn{1}{c|}{\rotatebox[origin=c]{-90}{{\begin{tabular}[c]{@{}c@{}}Blue \\ Box\end{tabular}}}} & \multicolumn{1}{l|}{\rotatebox[origin=c]{-90}{{\begin{tabular}[c]{@{}c@{}}B/W \\ QR-box\end{tabular}}}} & \multicolumn{1}{c}{\rotatebox[origin=c]{-90}{{\begin{tabular}[c]{@{}c@{}}Mean \\ Push\end{tabular}}}} & \multicolumn{1}{||c|}{\rotatebox[origin=c]{-90}{{
\begin{tabular}[c]{@{}c@{}}Mean\\\end{tabular}}}} \\ \hline
\setarstrut{\renewcommand*{\arraystretch}{0.2}}
\\
\restorearstrut
\cline{1-1}
\multicolumn{1}{|c|}{\multirow{2}{*}{{\textbf{\begin{tabular}[c]{@{}c@{}}Method\end{tabular}}}}} & \multicolumn{12}{c}{\multirow{2}{*}{{\textbf{Benign Condition}}}} & \multicolumn{1}{c}{} \\
\multicolumn{1}{|c|}{} & \multicolumn{12}{c}{} & \multicolumn{1}{c}{} \\ \hline
\multicolumn{1}{|c|}{\textbf{Just Encoder (\%)}} & \multicolumn{1}{c|}{20} & \multicolumn{1}{c|}{20} & \multicolumn{1}{c|}{0} & \multicolumn{1}{c|}{40} & \multicolumn{1}{c|}{0} & \multicolumn{1}{c|}{10} & \multicolumn{1}{c|}{15.0} & \multicolumn{1}{c|}{40} & \multicolumn{1}{c|}{10} & \multicolumn{1}{c|}{0} & \multicolumn{1}{c|}{0} & \multicolumn{1}{c}{12.5} & \multicolumn{1}{||c|}{14.0} \\ \hline
\multicolumn{1}{|c|}{\textbf{Traditional VAE (\%)}} & \multicolumn{1}{c|}{60} & \multicolumn{1}{c|}{60} & \multicolumn{1}{c|}{20} & \multicolumn{1}{c|}{20} & \multicolumn{1}{c|}{50} & \multicolumn{1}{c|}{30} & \multicolumn{1}{c|}{40.0} & \multicolumn{1}{c|}{50} & \multicolumn{1}{c|}{\textbf{60}} & \multicolumn{1}{c|}{\textbf{30}} & \multicolumn{1}{c|}{30} & \multicolumn{1}{c}{42.5} & \multicolumn{1}{||c|}{41.0} \\ \hline
\multicolumn{1}{|c|}{\textbf{\cite{rahmatizadeh2017vision}(w/o TFA) (\%)}} & \multicolumn{1}{c|}{70} & \multicolumn{1}{c|}{50} & \multicolumn{1}{c|}{30} & \multicolumn{1}{c|}{40} & \multicolumn{1}{c|}{60} & \multicolumn{1}{c|}{10} & \multicolumn{1}{c|}{43.3} & \multicolumn{1}{c|}{80} & \multicolumn{1}{c|}{\textbf{60}} & \multicolumn{1}{c|}{10} & \multicolumn{1}{c|}{20} & \multicolumn{1}{c}{42.5} & \multicolumn{1}{||c|}{43.0} \\ \hline
\multicolumn{1}{|c|}{\textbf{with TFA (\%)}} & \multicolumn{1}{c|}{\textbf{80}} & \multicolumn{1}{c|}{\textbf{80}} & \multicolumn{1}{c|}{\textbf{60}} & \multicolumn{1}{c|}{\textbf{50}} & \multicolumn{1}{c|}{\textbf{80}} & \multicolumn{1}{c|}{\textbf{40}} & \multicolumn{1}{c|}{\textbf{65.0}} & \multicolumn{1}{c|}{\textbf{100}} & \multicolumn{1}{c|}{\textbf{60}} & \multicolumn{1}{c|}{\textbf{30}} & \multicolumn{1}{c|}{\textbf{60}} & \multicolumn{1}{c}{\textbf{62.5}} & \multicolumn{1}{||c|}{\textbf{64.0}} \\ \hline
 & \multicolumn{12}{c}{\multirow{2}{*}{{\textbf{With Disturbance}}}} & \multicolumn{1}{c}{} \\
\multicolumn{1}{c}{} & \multicolumn{12}{c}{} & \multicolumn{1}{c}{} \\ \hline
\multicolumn{1}{|c|}{\textbf{\cite{rahmatizadeh2017vision} (w/o TFA) (\%)}} & \multicolumn{1}{c|}{10} & \multicolumn{1}{c|}{10} & \multicolumn{1}{c|}{0} & \multicolumn{1}{c|}{0} & \multicolumn{1}{c|}{0} & \multicolumn{1}{c|}{0} & \multicolumn{1}{c|}{3.3} & \multicolumn{1}{c|}{0} & \multicolumn{1}{c|}{30} & \multicolumn{1}{c|}{0} & \multicolumn{1}{c|}{0} & \multicolumn{1}{c}{7.5} & \multicolumn{1}{||c|}{5.0} \\ \hline
\multicolumn{1}{|c|}{\textbf{with TFA (\%)}} & \multicolumn{1}{c|}{\textbf{70}} & \multicolumn{1}{c|}{\textbf{80}} & \multicolumn{1}{c|}{\textbf{60}} & \multicolumn{1}{c|}{\textbf{60}} & \multicolumn{1}{c|}{\textbf{40}} & \multicolumn{1}{c|}{\textbf{40}} & \multicolumn{1}{c|}{\textbf{58.3}} & \multicolumn{1}{c|}{\textbf{90}} & \multicolumn{1}{c|}{\textbf{50}} & \multicolumn{1}{c|}{\textbf{30}} & \multicolumn{1}{c|}{\textbf{50}} & \multicolumn{1}{c}{\textbf{55.0}} & \multicolumn{1}{||c|}{\textbf{57.0}} \\ \hline
\end{tabular}
\end{adjustbox}
\vspace{0.1cm}
\caption{The upper half of the table shows the rate of successfully performing the desired manipulation with different sentence commands. The model with \textbf{TFA} has superior results  to a model without it \protect{\cite{rahmatizadeh2017vision}}. We also train a version of our model without the Discriminator, named \emph{Traditional VAE}. The model trained without $D$ cannot effectively perform the manipulations since the adversarial loss helps to learn rich Primary Latent Variable ($\mathbf{z}$). Also, in \emph{Just Encoder experiment}, we just use the Encoder as the visual network. The lower half of the table shows the rate of successfully performing the desired command, while being disturbed by an external agent. The model with \textbf{TFA} is by far better than a model without it \protect{\cite{rahmatizadeh2017vision}} in all cases.}
\label{tab:results}
\end{table*}

The first set of experiments studies the performance of the visuomotor controller under benign conditions, that is, under situations when the robot is given a textual command, $I_c$ in Sec.~\ref{sec:visualNetwork}, and it is left alone to perform the task in an undisturbed environment. To compare our approach against a baseline, we have reimplemented and trained the network described in~\cite{rahmatizadeh2017vision}, which can be used in the same experimental setup, but it does not feature a task focused visual attention. Note that the success rates are not directly comparable with~\cite{rahmatizadeh2017vision}, due to the more complex objects used here and the different camera position and environment of our robot. We trained the~\cite{rahmatizadeh2017vision} model on our own dataset, tuned its hyper-parameters and also tried to get the best possible results by adding all the loss terms explained in Sec.~\ref{sec:vismotnetworks}.

Table~\ref{tab:results} compares the performance of the four approaches for all the tasks, averaged over 10 tries each. We note that the proposed architecture using ``TFA'' outperforms the ``w/o TFA''  on all tasks. As an ablation study, we remove the discriminator and train the system as a traditional VAE (compared to VAE-GAN). Also, in another experiment we trained the E just by using the motor network loss without any GAN. We confirm the contribution of the adversarial loss and the GAN network to produce a rich primary latent variable $\mathbf{z}$. We observe that not having the adversarial loss will reduce the sharpness of the reconstructed images and fade out the details. Note that the model without adversarial loss fails to manipulate objects that require precise positioning like the black dumbbell or the blue ring, however, it can push the white plate much better as the plate is a big symmetric object. Please refer to the supplementary materials to compare the reconstructed images with and without the adversarial loss.

\subsection{Recovery after disturbance}

In the second series of experiments, we investigate the controller's ability to recover from a physical and visual disturbance. We are comparing the baseline model and our model which uses TFA. Physically disturbing means to disturbed the robot either by (a) pushing the object just when the robot was about to pick it up or (b) forcefully taking away the object from the robot after a successful grasp. For the push tasks, we bring in one or two hands into the scene (Figure~\ref{fig:gorilla_effect}). We make different visual disturbances by bringing in the hand in random positions, waving it, sometimes covering whole top part of the scene. In some cases we even put other random objects like a paper gorilla.

Under the described situations we count as success, if the robot notices the disturbance and recovers by successfully redoing the task. We remind the audience of the paper that due to the limitations of the Lynxmotion-AL5D robot, the {\em only} way the robot can detect the disturbance is through its visual system. 

Table~\ref{tab:results} shows the experimental results for scenarios with physical/visual disturbance. We notice that the results here are drastically better than the baseline. In the absence of TFA, the recovery rate is close to zero. In most cases, after loosing the object, the robot tried to execute the manipulation without noticing that it does not grasp the object. With the help of TFA, however, the robot almost always notices the disturbance, turns back and tries to redo the grasp. This phenomena is illustrated in our supplementary material video. Averaged over all the objects, the recovery rate is only 5\% for the baseline policy in pickup and push tasks, while it is \textbf{57\%} for the policy with the TFA (see Tables~\ref{tab:results}). Note that physical disturbance doesn't necessarily drop the robot's success rate since disturbing the robot occurs only when it is about to successfully perform the task, therefore the robot's success rate with and without the physical disturbance are not comparable. In other words, robot starts doing the task, a human judge decides if the robot is doing well and if it is, the human judge starts to disturbing the robot. We discard any tries that the robot is likely to fail even without disturbance.


\paragraph{The disappearing gorilla:} The proposed architecture allows us to ignore many of the possible visual disturbances. Experiments comparing the architecture to one without TFA confirm that this is indeed the case. Another way to study whether the policy ignores the visual disturbance is to reconnect the generator during test time as well, and study the reconstituted video frames (which are a good representation of the information content of primary latent encoding). Figure~\ref{fig:gorilla_effect} shows the input video frames (first row), the reconstructed video frames (second row) and the generated masked frames (third row). While the robot was executing the task of pushing the red bowl to the left, we added some disturbances such as waving a hand or inserting a cutout gorilla figure in the visual field of robot. Notice that in the reconstructed frames, the hand and the gorilla disappear, while the subject matter is reconstructed accurately. As these disturbing visual objects are ignored by the encoding, the task execution proceeds without disturbance. While we must be careful about making claims on the biological plausibility of the details of our architecture, we note that the overall effect implements a behavior similar to the selective attention experiments\footnote{\href{https://youtu.be/vJG698U2Mvo}{https://youtu.be/vJG698U2Mvo}} of Chabris and Simmons~\cite{chabris2010invisible}, purely as a side effect of an architecture implemented for a completely different goal. 

\section{Conclusion}
\label{sec:conclusion}

In this paper, we proposed a method for augmenting a deep visuomotor policy learned from demonstration with a task focused visual attention model. The attention is guided by a natural language description of the task -- it effectively tells the policy to ``Pay Attention!'' to the task and object at hand. Our experiments show that under benign situations, the resulting policy consistently outperforms a related baseline policy. More importantly, paying attention has significant robustness benefits. In severe adversarial situations, where a bump or human intervention forces the robot to miss the grasp or drop the object, we demonstrated through experiments that the proposed policy recovers quickly in the majority of cases, while the baseline policy almost never recovers. In the case of visual disturbances such as moving foreign objects in the visual field of the robot, the new policy is able to ignore these disturbances which in the baseline policy often trigger erratic behavior. 

Future work includes attention systems that can simultaneously focus on multiple objects, shift from object to object according to the requirements of the task, and work in severe clutter.
\\
\noindent{\bf Acknowledgments:} This work had been supported in part by the National Science Foundation under grant numbers IIS-1409823 and IIS-1741431. Any opinions, findings, and conclusions or recommendations expressed in this material are those of the authors and do not necessarily reflect the views of the National Science Foundation.
{\small
\bibliographystyle{IEEEtran}
\bibliography{ms}
}

\clearpage
\title{Supplementary Material: \\Pay attention! - Robustifying a Deep Visuomotor Policy through Task-Focused Visual Attention}
\author{Pooya Abolghasemi\samethanks, Amir Mazaheri\samethanks, Mubarak Shah and Ladislau B\"ol\"oni\\
University of Central Florida, Orlando, FL 32816\\
{\tt\small pooya.abolghasemi, amirmazaheri@knights.ucf.edu, shah@crcv.ucf.edu, lboloni@cs.ucf.edu}
}
\date{}

\maketitle
In the main manuscript, we propose a deep visuomotor policy that benefits from Task Focused visual Attention (TFA) and show that it outperforms policies using a task-independent visual network. More importantly, we show that the TFA has a very large impact when there are visual and physical disturbances in the environment.

In these supplementary materials we provide more details about the proposed architecture, and experimental settings. Moreover, we demonstrate some mid-level outputs of the proposed method as a comprehensive study. Additionally, we provide a video demo that includes some examples of our data collection process, experiments in benign condition, and experiments in presence of physical/visual disturbance.

\subsection*{Experimental Settings}
\begin{figure}
    \centering
    \includegraphics[width=0.6\columnwidth]{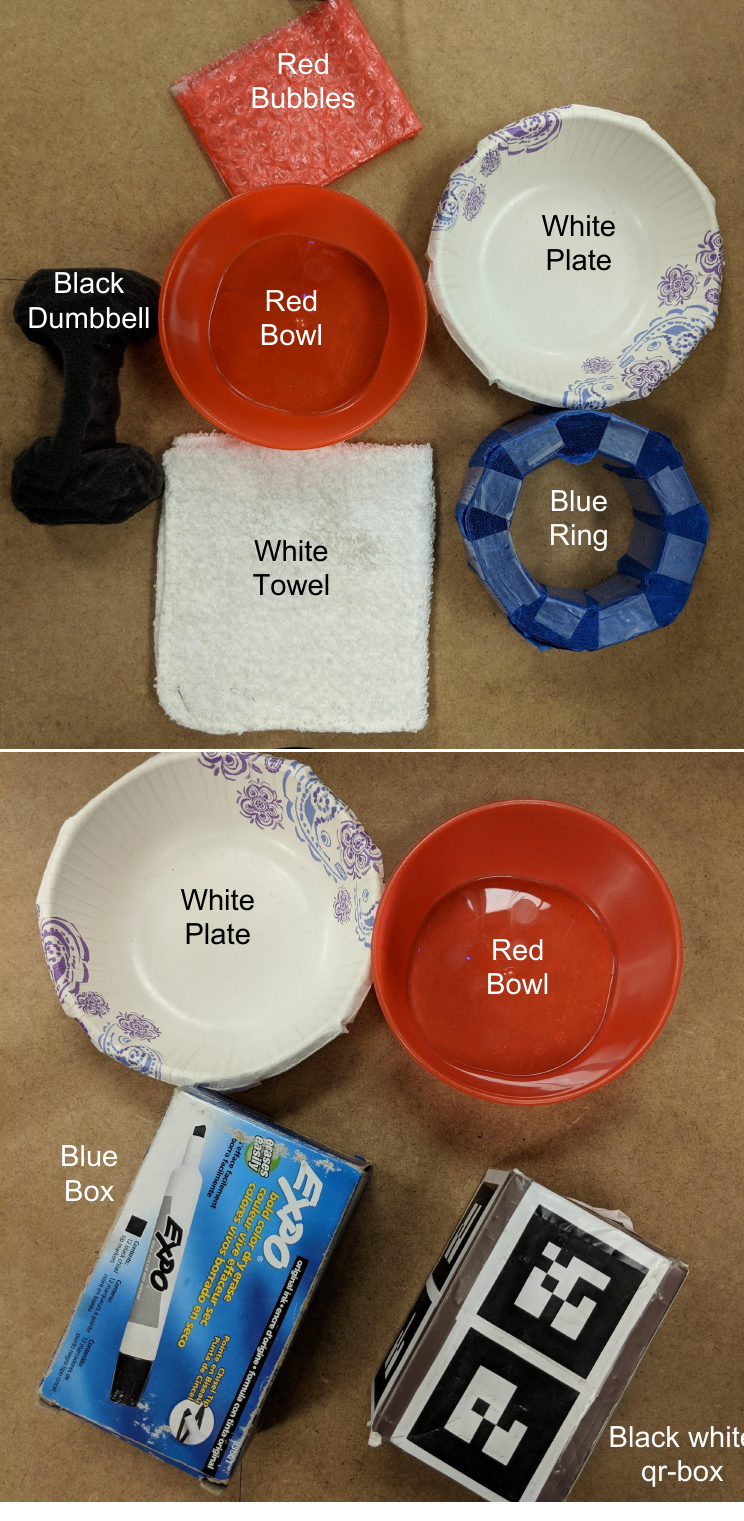}
    \caption{Objects used in all the picking up (top) and pushing (bottom) tasks experiments. (see Table 1 of the main manuscript) }
    \label{fig:demonstration objects}
\end{figure}

In this subsection, we provide additional details about our collected dataset and the experimental settings. Figure~\ref{fig:demonstration objects} shows all the objects used in our experiments listed in Table 1 of the main manuscript.

The experimental protocol we followed was as follows. The robot always starts from a fixed starting position and the task is considered successful if the robot performs the required task and returns back to the starting position. During testing, the robot has to finish the task within \emph{2 minutes}. A human judge observes the robot during the performance and decides if it has been successful or not.

For the experiments with physical/visual disturbance, the human judge decides if the robot is likely to succeed to perform the task and then provides the disturbance. We stop and do not consider experiments where the robot is clearly failing even without the disturbance (for example, if it doesn't get close to the object correctly or accidentally push the object to an out of access point before the disturbance starts). Thus, our measurements on the recovery from disturbance, only consider the cases when the need for recovery was caused by the disturbance.
\subsection*{Teacher Network Details}
In Section 3.1 of the main manuscript, we describe the Teacher Network for visual attention. The masked frames produced by the teacher network are considered as ``real'' masked frames (denoted by $m$). Also, the teacher network is pre-trained separately (has its own loss function and optimizer), and helps the proposed network to produce the Primary Latent Variable $z$ that is a rich representation of the visual world and is robust to disturbance. We consider the visual attention produced by the teacher network as ground truth attention.

In Table~\ref{tab:teacherDetails} we show the hyper-parameter values used in our implementation of the teacher network. 

\begin{table}
    \centering
    \begin{tabular}{|c|c|}
        \hline
        Hyper-parameters & Value \\
        \hline
        $\|V\|$ & 20 \\
        $d_v$ & 200 \\
        $d_h$ & 200 \\
        $k$ & 196 \\
        $d_\phi$ & 512 \\
        $d_\psi$ & 200 \\
        \hline
    \end{tabular}
    \caption{Hyper-parameter values in our implementation of teacher network, described in the Section 3.1 of the main manuscript. }
    \label{tab:teacherDetails}
\end{table}

\begin{figure*}
    \centering
    \includegraphics{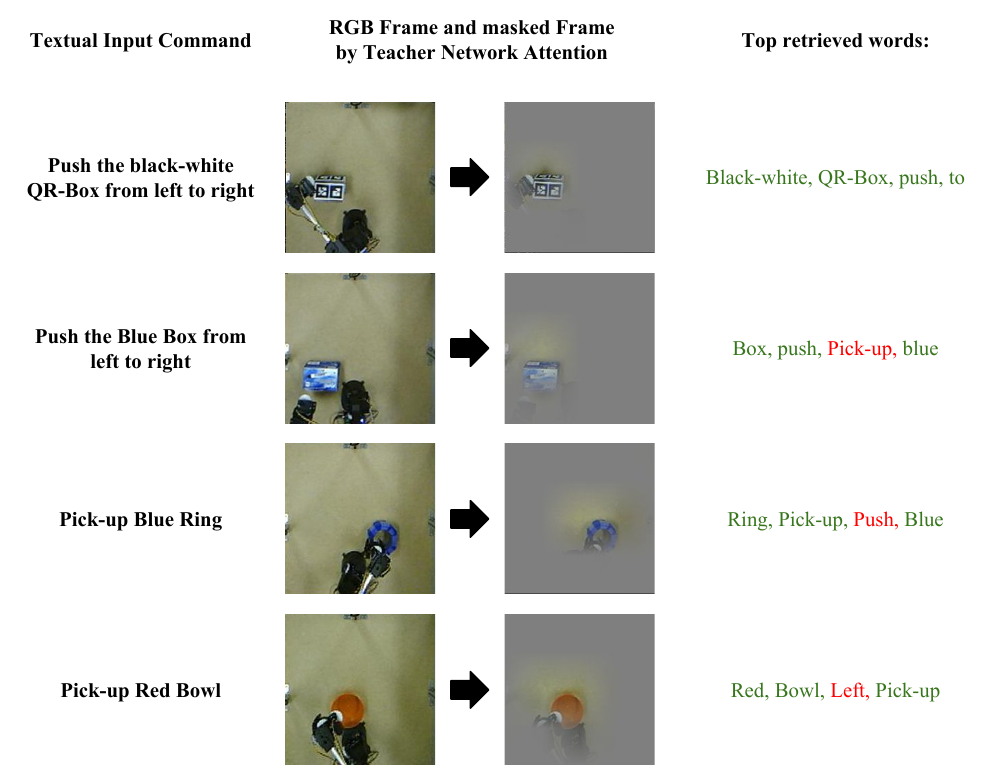}
    \caption{Here we show how well the trained teacher network can select correct words about a task being performed based on the generated visual attention. We show the textual input command, one frame, and the masked frame using the attention computed by the teacher network, and the top retrieved words by teacher network. Basically, we sort the scores $\hat{\mathcal{V}}$ in Equation 5 of the main manuscript, and show the top 4 words. The green/red color indicates the words which are/aren't in common with the textual input command.}
    \label{fig:TeacherExamples}
\end{figure*}

As explained in Section 3.1, we pre-train the teacher network by reconstructing the set of words used in the textual input sentence (refer to the $\mathcal{L}_{att}$ loss in the main manuscript). Figure~\ref{fig:TeacherExamples} demonstrates a few examples of the trained teacher network and shows how good it can select the words appeared in the input sentence just based on the visual features of the attended spatial regions (Equation 5 of the main manuscript). We observe that it can predict all the object names and colors correctly. In some cases, since the input is only one frame, it is impossible to predict the verb words (push or pickup in our experiments). For example, in top-right example of Figure~\ref{fig:TeacherExamples}, since the robot arm has not reached the object, it is not easy to say if it going to be a push or pick-up task. On the other hand, in the top-left example, the robot arm has reached next to the object and it is clear that it is going to push it, and the teacher network also correctly predict the verb ``push''.

\subsection*{Detailed Architectures}
\begin{figure*}[t]
    \centering
    \includegraphics[width=\textwidth]{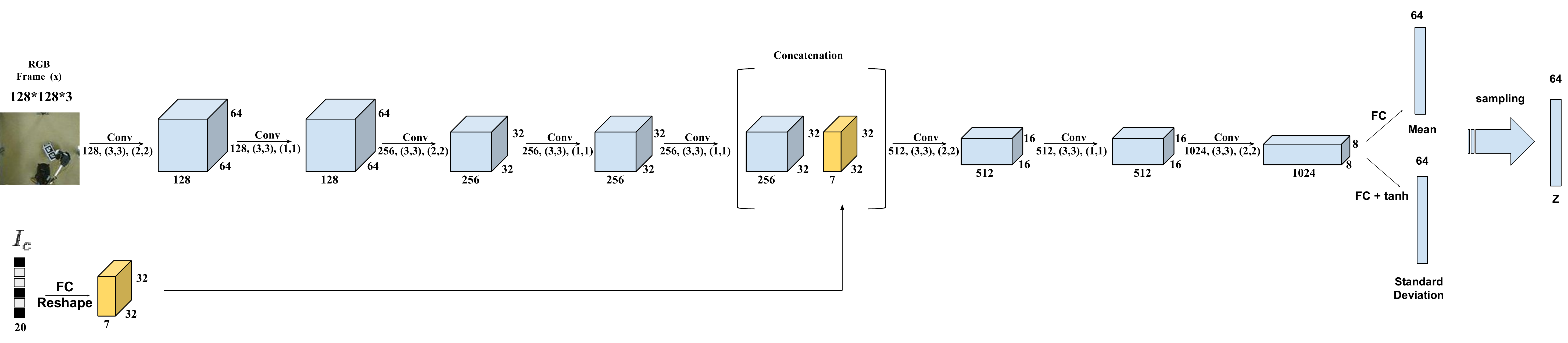}
    \caption{Encoder Architecture used in our framework. The textual command $I_c$ and  RGB frame are the inputs to the Encoder. The Primary Latent Variable $z$ is the output of the Encoder. Here, we show all the layers used in the Encoder including Convolution and Fully-Connected (FC) layers. We also indicate the number of filters, kernel size, and the stride of each convolution. Note that, in our implementation, all the convolutions layers are followed by Batch-Normalization and a Leaky Relu.}
    \label{fig:Encoder}
\end{figure*}

\begin{figure*}[t]
    \centering
    \includegraphics[width=\textwidth]{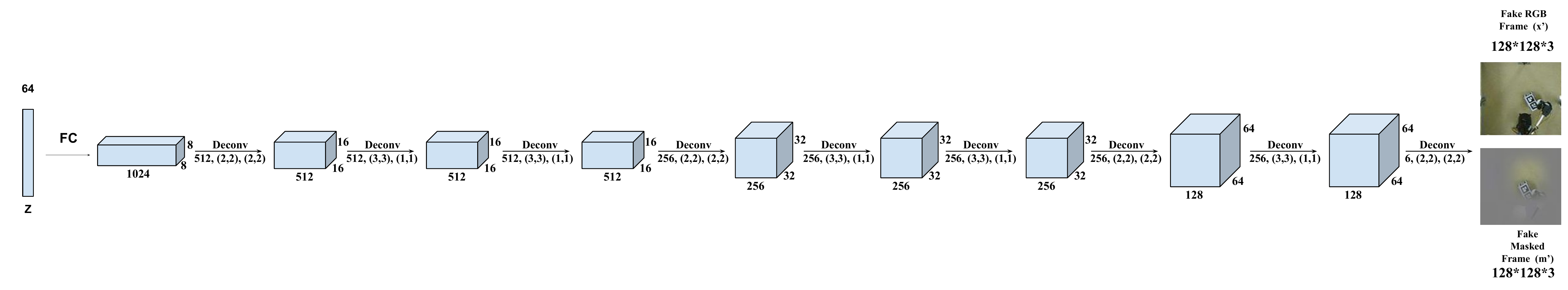}
    \caption{Generator Architecture used in our framework. The input to the generator is the Primary Latent Variable $z$, which is produced by the Encoder (Figure~\ref{fig:Encoder}). And the outputs of generator are two images, a fake frame ($x'$) and a fake masked frame ($m'$) respectively. We use multiple layers of Deconvolutions to generate an image out of the input vector. We show the number of filters, kernel size and the stride of each Deconvolution. All the Deconvolutions are followed by Batch-Normalization and Relu.}
    \label{fig:Generator}
\end{figure*}

\begin{figure*}[t]
    \centering
    \includegraphics[width=\textwidth]{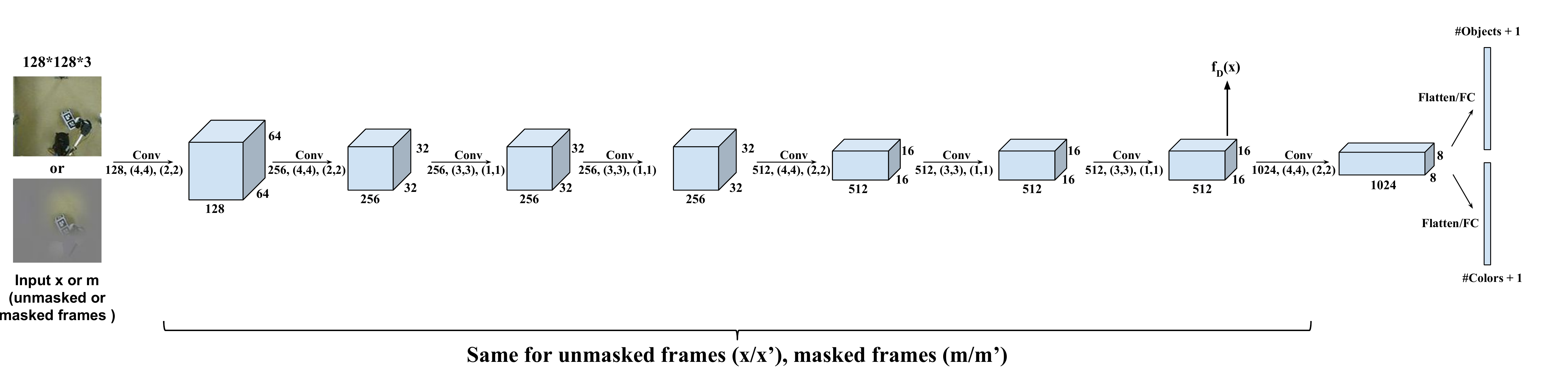}
    \caption{Discriminator Architecture used in our framework. Our proposed discriminator architecture not only distinguishes between the fake and real frames, but also classifies the object type and the color of the object being manipulated. Convolution layers are shared for both types of inputs (masked or unmasked) but the last Fully Connected (FC) layers are separate for each kind of input.}
    \label{fig:Discriminator}
\end{figure*}

\begin{figure*}[t]
    \centering
    \includegraphics[width=\textwidth]{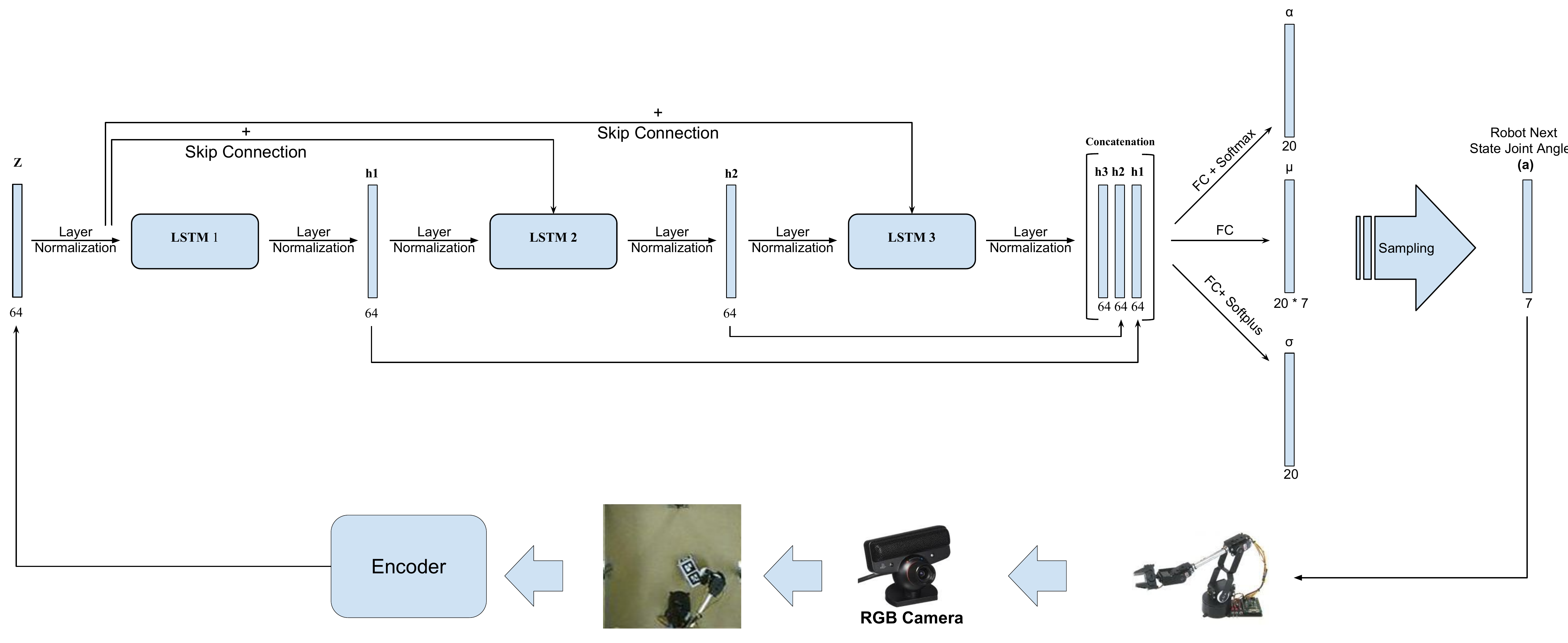}
    \caption{Motor Network Architecture used in our framework. Given the Primary Latent Variable $z$, the motor network predicts the next state of the robot, and produces 7 numbers (corresponding to 7 joint angles of the robot) to move the joints of the robot. We use 3 stacked layers of LSTMs with skip connections. Also, we use layer normalization in between LSTMs. We concatenate the outputs of all the LSTMs and generate the $\mu$, $\sigma$, and the mix coefficients $\alpha$ for the Mixture Density Network (MDN) described in the main manuscript. We use 20 Gaussians for the MDN of our implementation. The states of all LSTMs get updated frame by frame. In fact, after each move of the robot, a new frame is captured by the RGB camera, fed to the Encoder, and then the next $z$ vector is fed back to the Motor Network. Each frame corresponds to one time-step for LSTMs.}
    \label{fig:motornetwork}
\end{figure*}

In this section, as promised in the main manuscript, we show the detailed architectures of our Encoder (Figure~\ref{fig:Encoder}), Generator (Figure~\ref{fig:Generator}), Discriminator (Figure~\ref{fig:Discriminator}) and Motor network (Figure~\ref{fig:motornetwork}).

Note that, except the visual Teacher Network which is trained separately, all other modules including Encoder, Generator, Discriminator, and Motor Network are trained end-to-end.
\subsection*{Attention and Disturbance}
In the main manuscript, we show that our proposed approach with Task-Focused visual Attention (TFA) makes the robot policy robust to various types of visual and physical disturbances. Table 1 of the main manuscript shows that the average performance of the model without TFA is about 50\% less than the proposed network. Here, by visualizing the generated images from the two named experiments, we qualitatively show the reasons behind robustness and effectiveness of visual attention during the disturbance.

\begin{figure*}[ht]
    \centering
    \includegraphics[width=\textwidth]{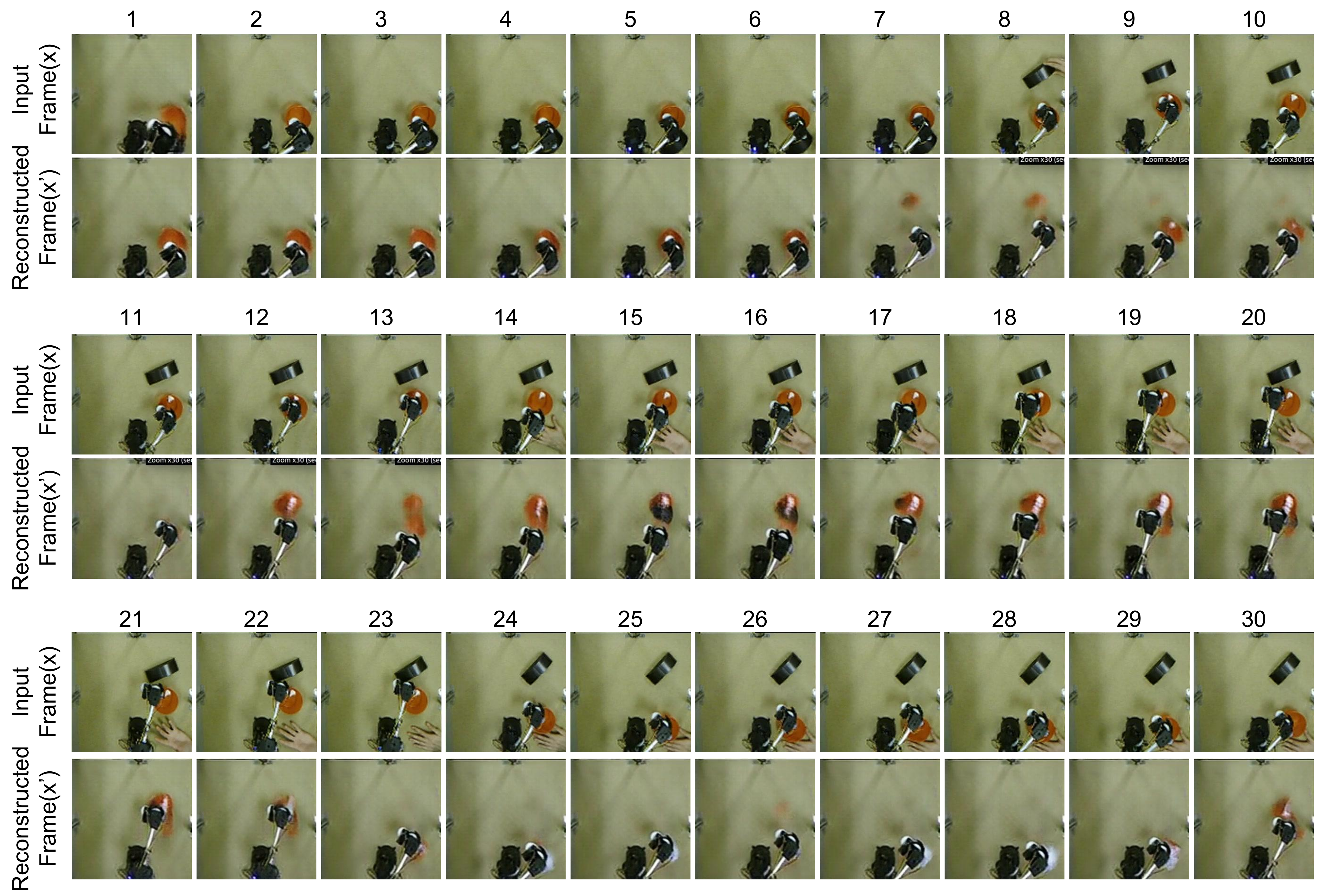}
    \caption{A sequence of frames from an experiment with visual disturbance. The textual command for this experiment is: "push the red bowl from left to right". Here, we show the frame reconstructions of the ``w/o TFA'' model. In many frames like 23-29, the model has failed to reconstruct the input frame properly, showing that the Primary Latent Variable $z$ in this model is not robust to visual disturbance.}
    \label{fig:roohi_obj_dist}
\end{figure*}

Figure~\ref{fig:roohi_obj_dist} shows the original and reconstructed frames from the ``w/o TFA'' experiment. The frames show the sequence of a task when two other objects, namely a human hand and an eyeglass box, enter the scene and disturb the internal representation (starting frame 8). We notice that the reconstructed frames become very blurry and inaccurate. For example, from frame 23 to 30, the object of interest is completely missing in the reconstruction. We conjecture that the disturbance forced the primary latent encoding $z$ to move to an unseen state from which the generator cannot reconstruct meaningful images. 

\begin{figure*}[t]
    \centering
    \includegraphics[width=\textwidth]{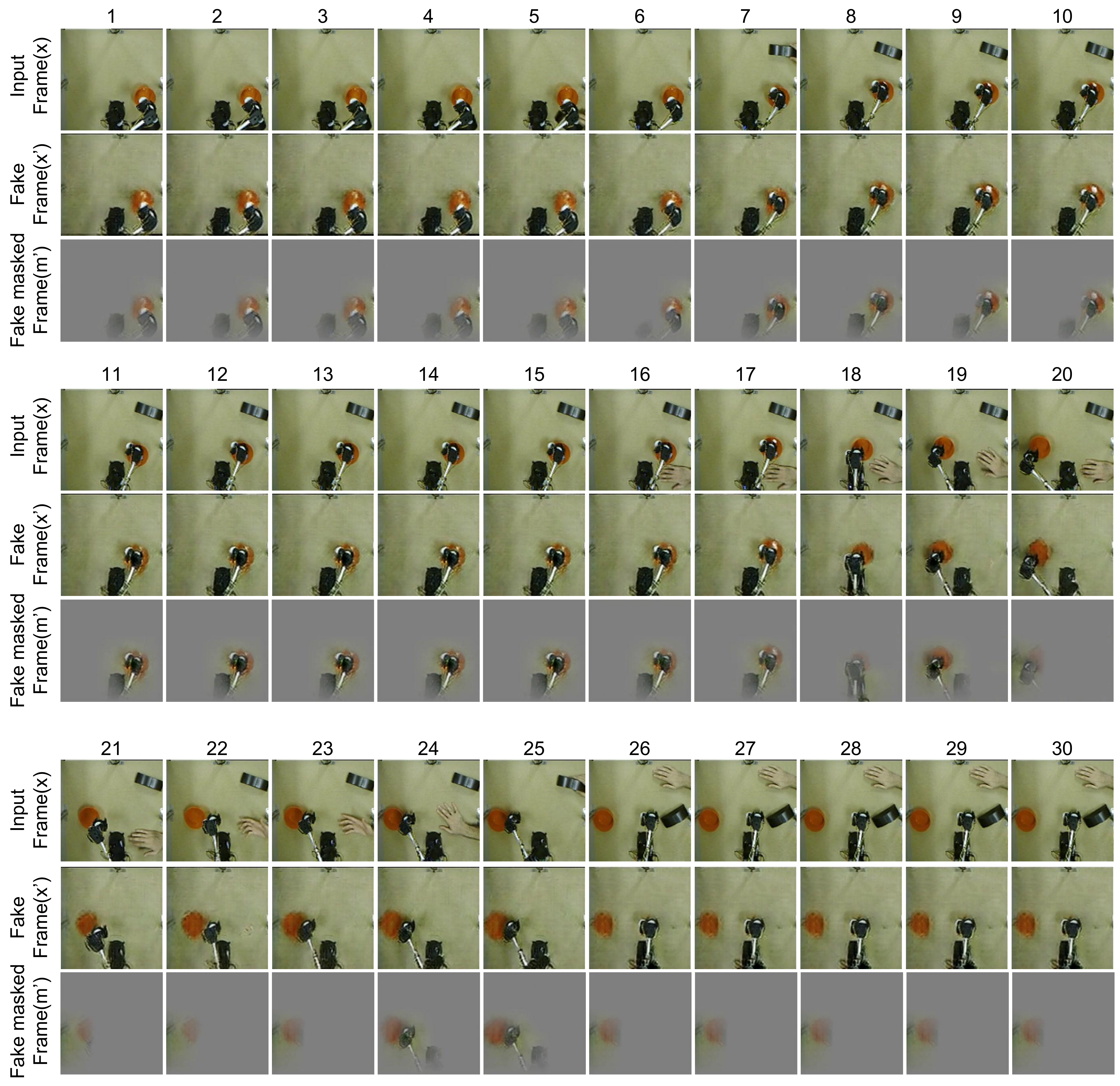}
    \caption{A sequence of frames of an example with a scenario similar to the one in Figure~\ref{fig:roohi_obj_dist}, using the TFA-augmented model. We notice that the attention stays on the correct object and the model has a better reconstruction of the object in both of masked and unmasked frames compared to Figure~\ref{fig:roohi_obj_dist}.}
    \label{fig:tfa_obj_dist}
\end{figure*}

Figure~\ref{fig:tfa_obj_dist} reproduces same disturbance scenario as in Figure~\ref{fig:roohi_obj_dist} but using the model with TFA. We notice that the attention disregards the obstacles and disturbances and the quality of reconstructed frames do not drop drastically. Also, we see that the disturbing objects such as the hand are removed from reconstructions. 


\subsection*{Attention Examples}

\begin{figure*}[ht]
    \centering
    \includegraphics[width=0.9\textwidth]{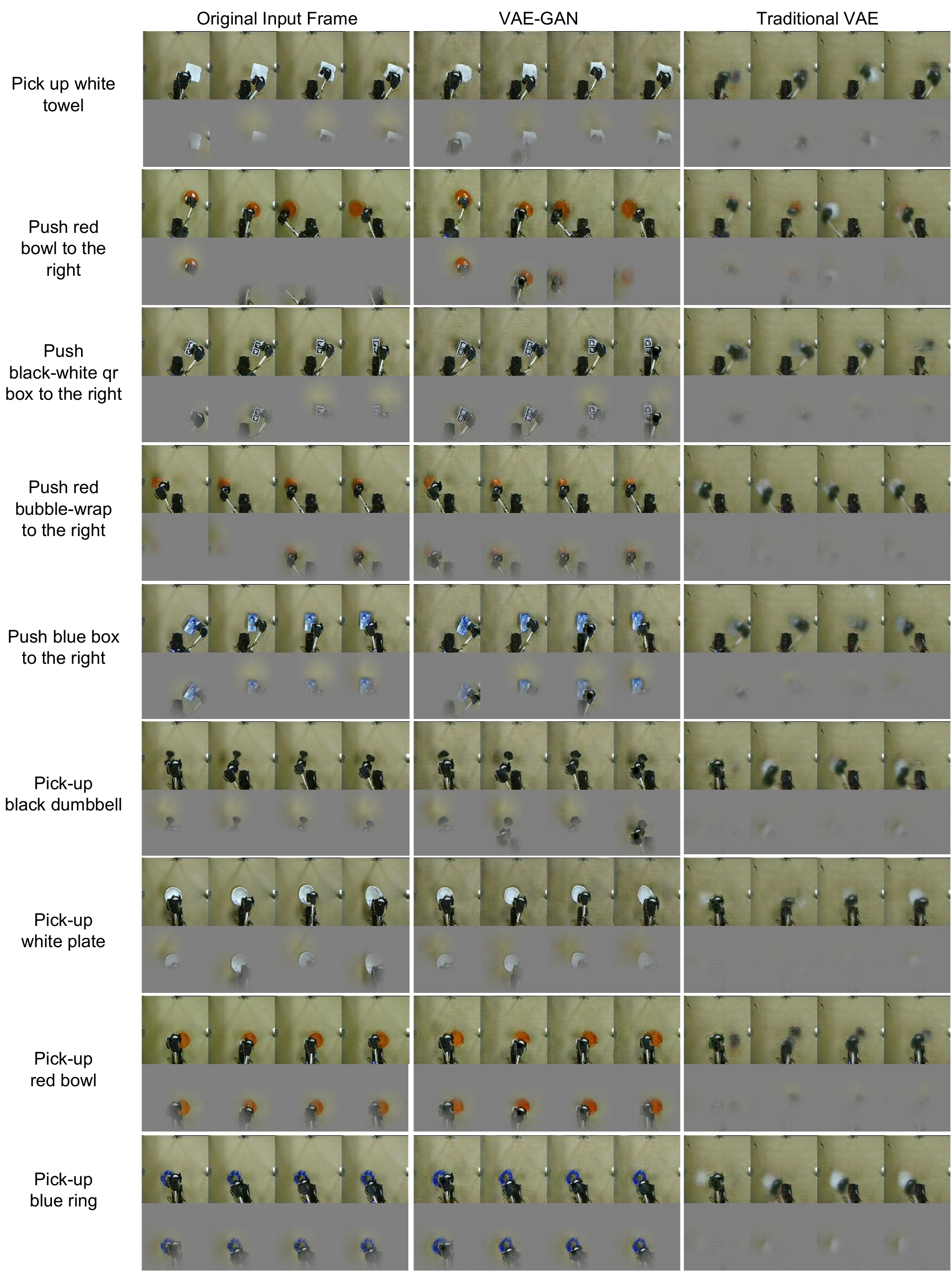}
    \caption{A comparison between the original frame, reconstructed frames by VAE-GAN and Traditional VAE. We also show the real masked frames by attention (using the teacher network), generated fake masked frame by VAE-GAN and Traditional GAN. By comparing the second and third columns of this figure, we can justify the performance drop of the ``Traditional VAE'' experiment in Table 1 of the main manuscript.}
    \label{fig:vaevsvaegan}
\end{figure*}

Figure~\ref{fig:vaevsvaegan} illustrates several frames for a given textual command in each row. The first column of each row shows the original frames (denoted by $x$ in the main manuscript), and the masked frames produced by the teacher network (denoted by $m$ in the main manuscript). The second and third columns show the fake (reconstructed) frame and the fake masked frame generated by the generator ($x'$ and $m'$ in the main manuscript); however, the third column shows the results out of generator when it is trained merely based on a reconstruction loss, without any discriminator. The quality difference between the second and third columns reconstructions explains the performance difference between ``traditional VAE'' and other experiments with the ``VAE-GAN'' setting (see Table 1 of the main manuscript).

We notice that in some cases, the attention produced by the generator is even better than the teacher network attention. For example, compare the masked frames of the first and second columns of the second and fourth examples in Figure~\ref{fig:vaevsvaegan}. We believe that this phenomena is due to the rich Primary Latent Variable, $z$ that our network learns. In fact, the fake masked frame must be rich enough that the discriminator predict the correct object and color of the task and it provides some complementary information to the Encoder.

\end{document}